\documentclass[10pt,a4]{article}
\usepackage[top=1.0cm,left=2.0cm,footskip=0.5cm,marginparwidth=1cm]{geometry}
\usepackage{graphicx}
\usepackage{amsmath}
\usepackage{algorithm}
\usepackage{algorithmicx}
\usepackage{algpseudocode}
\usepackage{cite}
\usepackage{url}
\usepackage{textcomp}
\usepackage{xcolor}
\usepackage{hyperref}
\usepackage{subfig}
\usepackage{listings}

\usepackage[utf8]{inputenc}

\usepackage{cite}

\usepackage{nameref,hyperref}

\usepackage[right]{lineno}

\usepackage{microtype}
\DisableLigatures[f]{encoding = *, family = * }

\usepackage{changepage}

\usepackage[aboveskip=1pt,labelfont=bf,labelsep=period,singlelinecheck=off]{caption}

\makeatletter
\renewcommand{\@biblabel}[1]{\quad#1.}
\makeatother

\usepackage{color}

\definecolor{Gray}{gray}{.25}

\usepackage{graphicx}

\usepackage{sidecap}

\usepackage{wrapfig}
\usepackage[pscoord]{eso-pic}
\usepackage[fulladjust]{marginnote}
\reversemarginpar

\newcommand\YAMLcolonstyle{\color{red}\mdseries}
\newcommand\YAMLkeystyle{\color{black}\bfseries}
\newcommand\YAMLvaluestyle{\color{blue}\mdseries}
\makeatletter
\newcommand\language@yaml{yaml}
\expandafter\expandafter\expandafter\lstdefinelanguage
\expandafter{\language@yaml}
{
  keywords={true,false,null,y,n},
  keywordstyle=\color{darkgray}\bfseries,
  basicstyle=\YAMLkeystyle,                                 
  sensitive=false,
  comment=[l]{\#},
  morecomment=[s]{/*}{*/},
  commentstyle=\color{purple}\ttfamily,
  stringstyle=\YAMLvaluestyle\ttfamily,
  moredelim=[l][\color{orange}]{\&},
  moredelim=[l][\color{magenta}]{*},
  moredelim=**[il][\YAMLcolonstyle{:}\YAMLvaluestyle]{:},   
  morestring=[b]',
  morestring=[b]",
  literate =    {---}{{\ProcessThreeDashes}}3
                {>}{{\textcolor{red}\textgreater}}1     
                {|}{{\textcolor{red}\textbar}}1 
                {\ -\ }{{\mdseries\ -\ }}3,
}
\lst@AddToHook{EveryLine}{\ifx\lst@language\language@yaml\YAMLkeystyle\fi}
\makeatother
\newcommand\ProcessThreeDashes{\llap{\color{cyan}\mdseries-{-}-}}

\begin{document}

\begin{flushleft}
{\Large
\textbf\newline{Pogosim -- a Simulator for Pogobot robots}
}
\newline
\\
Leo Cazenille\textsuperscript{1,*},
Loona Macabre\textsuperscript{1},
Nicolas Bredeche\textsuperscript{1,*}
\\
\bigskip
\bf{1} Sorbonne Universit\'{e}, CNRS, ISIR, F-75005 Paris, France
\\
\bigskip
* correspondence: leo.cazenille@gmail.com,nicolas.bredeche@sorbonne-universite.fr

\end{flushleft}


\begin{abstract}
Pogobots are a new type of open-source/open-hardware robots specifically designed for swarm robotics research. Their cost-effective and modular design, complemented by vibration-based and wheel-based locomotion, fast infrared communication and extensive software architecture facilitate the implementation of swarm intelligence algorithms. However, testing even simple distributed algorithms directly on robots is particularly labor-intensive. Scaling to more complex problems or calibrate user code parameters will have a prohibitively high strain on available resources. In this article we present Pogosim, a fast and scalable simulator for Pogobots, designed to reduce as much as possible algorithm development costs. The exact same code will be used in both simulation and to experimentally drive real robots. This article details the software architecture of Pogosim, explain how to write configuration files and user programs and how simulations approximate or differ from experiments. We describe how a large set of simulations can be launched in parallel, how to retrieve and analyze the simulation results, and how to optimize user code parameters using optimization algorithms. 

\end{abstract}







\section{Introduction}

Swarm robotics~\cite{hamann2018swarm} studies how large groups of relatively simple robots, each following local rules and exchanging only short-range information, can self-organize into robust, scalable, and flexible collectives that accomplish tasks no single robot could achieve alone. This emerging field takes inspiration from social insects and complex systems whose global behavior emerges from local interactions under constraints on sensing, communication, and actuation~\cite{csahin2004swarm}.
Landmark demonstrations—such as a thousand-robot self-assembly~\cite{rubenstein2014programmable} show the feasibility and challenges of bringing these principles to physical swarms.

\begin{figure}[htbp]
    \centering
    \resizebox{0.90\textwidth}{!}{%
    \includegraphics[height=3cm]{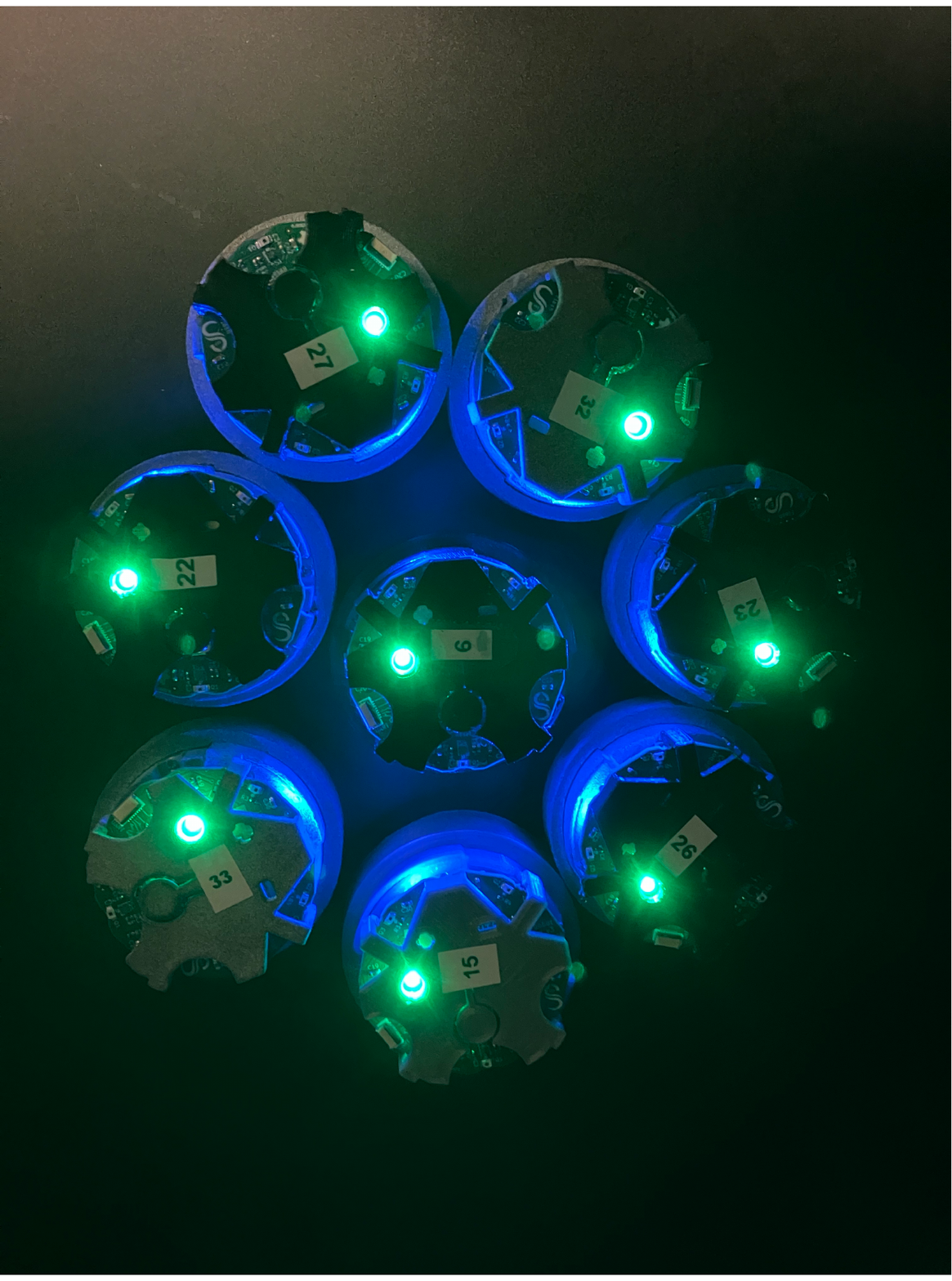}
    \hspace{0.1cm}%
    \includegraphics[height=3cm]{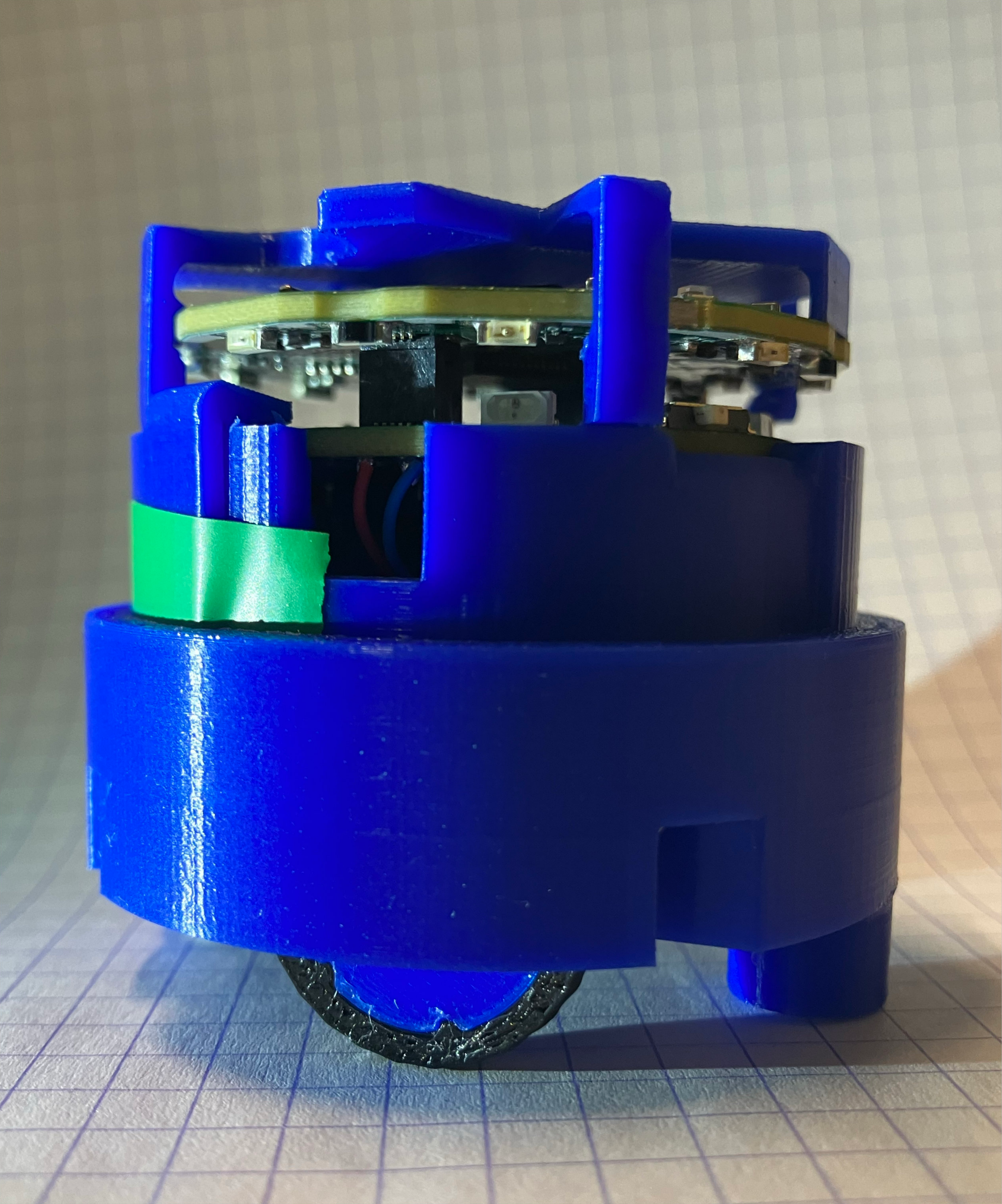}%
    \hspace{0.1cm}%
    \includegraphics[height=3cm]{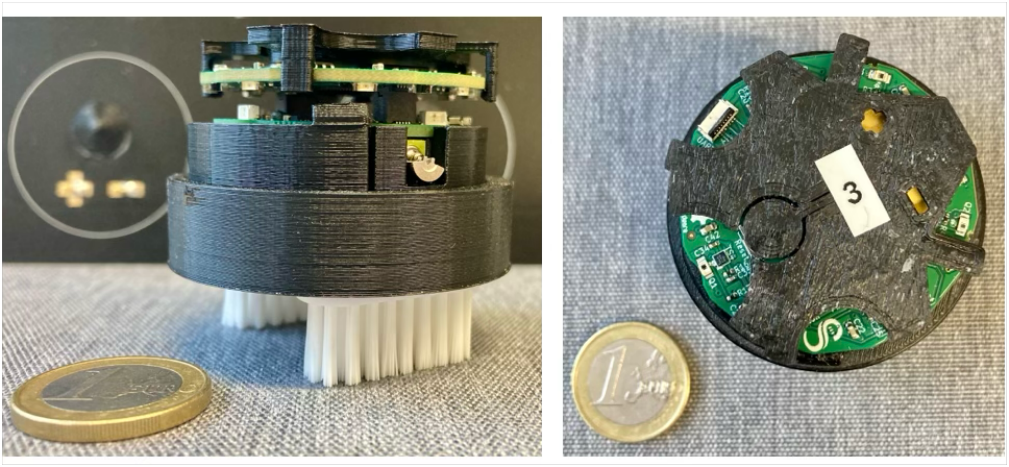}%
    }
    \caption{(1) a small swarm of Pogobot; (2) Pogobot with wheel-based locomotion; (3) Pogobot with vibration-induced locomotion, using toothbrush head with inclined brush; (4) view from above (the Pogobot is approx.~6~cm diameter).}
    \label{fig:pogobots}
\end{figure}

Pogobots~\cite{loi2025pobogot}, as illustrated in Fig.~\ref{fig:pogobots}, are open-source, open-hardware robots purpose-built for swarm-robotics research. They are compact (~6 cm diameter), low-cost, and modular, with interchangeable wheel- and vibration-based locomotion, fast very directional infrared (IR) communication, and a pragmatic software stack that targets large-scale experiments. The platform’s design (RISC-V softcore on FPGA, C API, and 3D-printable form factors) emphasizes accessibility without sacrificing capabilities, enabling studies that range from programmable active matter to distributed online learning. More than $200$ Pogobots are already being used on a daily basis at Sorbonne Université and PSL to study self-organizing systems, programmable active matter, discrete reaction-diffusion-advection systems as well as models of social learning and evolution.

However, developing and validating collective behaviors directly on hardware is costly, slow, and difficult to reproduce at scale—an issue long recognized in the Kilobot~\cite{rubenstein2012kilobot} (an older type of swarm robots) community and a central motivation for the Kilombo~\cite{jansson2015kilombo} simulator for Kilobots. Simulators allow rapid iteration, parameter sweeps, stress-tests under controlled noise, and fair comparisons across algorithms before committing lab time and wear-and-tear to robots. Crucially, when the simulator executes the very same controller code as the real robots, it minimizes “sim-to-real” drift and avoids duplicated code paths.

We introduce Pogosim, a fast, scalable simulator that mirrors Pogobot hardware and APIs so the exact same C controllers run unmodified in simulation and on real robots. Built on a solid-body physics engine (Box2D) with pseudo-realistic IR communication (including collision/loss models), Pogosim adds a lightweight GUI (SDL2), YAML-based scenario configuration, and additional tooling for batch experiments and offline parameter search (e.g. CMA-ES, MAP-Elites), with results serialized to compact Apache Arrow format for downstream analysis. Together, these components make it easy to launch thousands of runs, evaluate task performance and emergent dynamics, and carry the best controllers directly to hardware -- closing the loop between design, simulation, and experiment.

\begin{figure}[htbp]
    \centering
    \includegraphics[height=12.0cm]{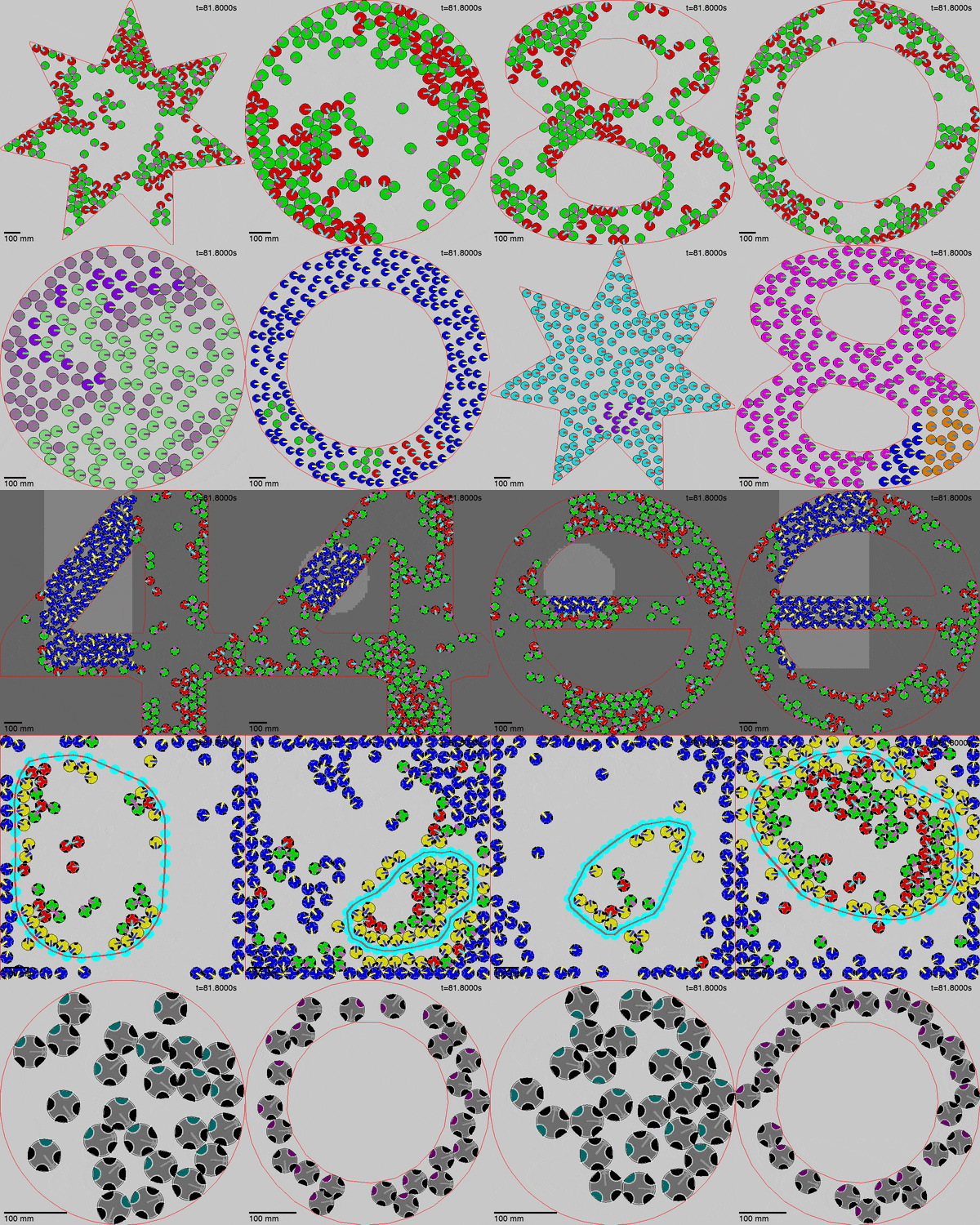}%
    \caption{Gallery of simulated runs of representative examples of Pogobots swarm dynamics simulated by Pogosim (C codes available at \url{https://github.com/Adacoma/pogosim/tree/main/examples}) and across several arenas. The top row showcases the ''run\_and\_tumble`` example (red robots: tumble, green: run). The 2nd row shows the ``hanabi'' example (diffusion of a color across the swarm and consensus). The 3rd row presents the ``phototaxis'' example (robots stop when light levels are high enough). The 4th row shows different behaviors close to the active walls (robots stop) and a soft moving active membrane. The 5th row displays results across collective shape classification (cyan LED=disk detected, violet=annulus detected) using the spectral swarm robotics (SSR) algorithm~\cite{cazenille2024hearing}.}
    \label{fig:gallery}
\end{figure}

Figure~\ref{fig:gallery} showcases the capabilities of Pogosim to scale to diverse and complex controllers, and to large number of robots and objects.

\section{Design}

\subsection{Design choices}\label{sec:design}
We designed Pogosim so that the exact same C code could be used both for simulation and experiments (robot controllers), without minimal differences between simulations and actual robotic experiments. This is achieved by re-implementing parts of the Pogobot internal API, and simulating base behavior of the robot hardware, communication system and physical dynamics (collisions, obstacles, friction, motor biases and noise, etc).

\begin{figure}[ht]
    \centering
    \includegraphics[width=1.0\textwidth]{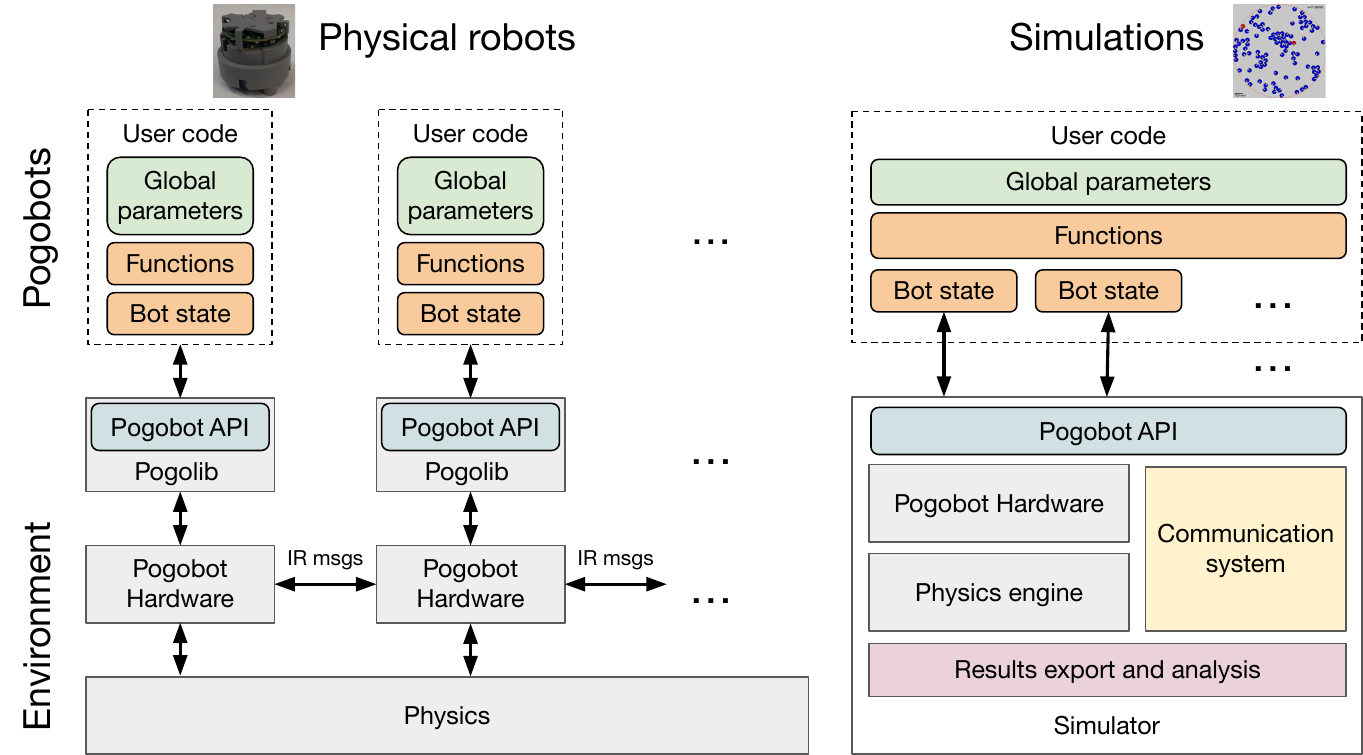}%
    \caption{Workflow of the simulator components (right panel) compared to experimental setups with physical Pogobots (left panel). In the latter, each robot has its own hardware and run its own instance of the user program and Pogolib. In Pogosim, all simulated robots share the same user program, and global parameters, but different and independent local states (local variables). The simulator re-implements the Pogolib API so that the simulated robots have access to the exact same sets of hardware items. Physical interactions between robots and other objects in the environment (e.g. collisions) are handled through a physics engine (Box2D~\cite{box2d}). The IR communication between robots is also simulated by Pogosim with a pseudo-realistic system (incl. communication loss).}
    \label{fig:workflow}
\end{figure}

Figure~\ref{fig:workflow} compares experimental setups composed of physical Pogobots with the Pogosim implementation and mappings. The simulated robots share the same C user code and the same global parameters, however robot instance has its own local parameters, saved in a dedicated structure. Using compiler optimization, local elements of this structure will be transformed into true globals when compiling binaries for physical Pogobots, avoiding increased computational costs.
The simulator re-implements the entire Pogolib API, including motor control, sensing, and message I/O. Physical dynamics between robots are handled by the Box2D solid-bodies physical engine~\cite{box2d}.

Pogosim processes simulation time with varying levels of realism depending on the component. It will be very precise for motor control and message I/O. However, it does not realistically model the amount of time spent executing user code. Moreover, Pogosim does not take into account the limited computational costs available on Pogobot FPGA/CPUs. As such, it is possible to create an user-code that will perform a large number of complex computations and API calls and make it work on simulations, but would become far slower on physical Pogobots.

Pogosim has a native support for several platforms: Linux, MacOSX and WSL2. We focused on those platforms because they are the three most prevalent ones in research laboratories.

\subsection{Compiling as a simulation or as a Pogobot binary}
Using Pogosim build system, it is possible to compile the same code into a simulation or a Pogobot binary, with the following commands:
\begin{lstlisting}[basicstyle=\ttfamily,language=bash]
make clean sim  # To compile the simulation
# OR
make clean bin  # To compile the binary for real Pogobots
# OR
make clean all  # To compile both the simulation and Pogobot binaries
\end{lstlisting}

By default, the name of the created simulation binary corresponds to the name of the parent directory of the project. You can then launch it using:
\begin{lstlisting}[basicstyle=\ttfamily,language=bash]
make clean sim
./template_prj -c conf/test.yaml        # If the parent directory is "template_prj"
\end{lstlisting}

You can rely on the same methods as baseline Pogobots to upload the binary on physical robots. E.g. on a Pogobot connected with an USB cable:
\begin{lstlisting}[basicstyle=\ttfamily,language=bash]
make connect TTY=/dev/ttyUSB0
serialboot
\end{lstlisting}

\subsection{Simulation configuration and instantiation}
Simulations can be parametrized through a YAML configuration file. The YAML format was selected because it has become one of the most well-spread configuration file format.
These configuration files contain information about the spatio-temporal properties of the setup (e.g. arena shape and surface, duration of the simulation), the simulated robots (number, physical parameters like shape, radius, weight, friction, damping, messaging capability, etc), and values of global parameters in the user code (shared by all robots).

Here is an example of a Pogosim YAML configuration file:
\begin{lstlisting}[language=yaml,basicstyle=\ttfamily\scriptsize]
---
window_width: 600       # In pixels
window_height: 600      # In pixels

arena_file: arenas/disk.csv
arena_surface: 1.0e6    # In mm^2

delete_old_files: true
enable_data_logging: true
data_filename: "frames/data.feather"
enable_console_logging: true
console_filename: "frames/console.txt"
save_data_period: 1.0       # In s, or -1 to disable data export (see also "enable_data_logging")
save_video_period: 1.0      # In s, or -1 to disable frame export
frames_name: "frames/f{:010.4f}.png"

seed: 0

# In the GUI, can enabled/disabled using F5
show_communication_channels: false
# Flag indicating if the communication channels must be drawn above the objects (true) or below.
show_communication_channels_above_all: false
# In the GUI, can enabled/disabled using F6
show_lateral_LEDs: false
# In the GUI, can enabled/disabled using F6
show_light_levels: false
# Enable/disable GUI
GUI: true

# Time
simulation_time: 50.0    # In s
time_step: 0.01            # In s
GUI_speed_up: 10.0         # How much the visualisation in the GUI should be sped up

# Describe the initial pose of the robots in the arena
initial_formation: random

# List of objects created in the simulation, by category
objects:
    # An object category containing a single light spot encompassing the entire simulation
    global_light:
        # The light is not active (does not move and does not have a controller)
        type: static_light
        # Type of light level paradigm: static (same value across the entire geometry) or gradient
        light_mode: static
        # The light involve the entire simulation
        geometry: global
        # Value of the light between 0 and 32767, detected by the Pogobots (pogobot_photosensors_read function)
        value: 200

    # An object category containing the Pogobot robots
    robots:
        type: pogobot       # Category type pertaining to Pogobots
        nb: 100             # Number of objects (Pogobots) in this category
        geometry: disk                  # Pogobots are always disk-shaped
        radius: 26.5                    # In mm

        # Physical properties
        body_linear_damping: 0.3
        body_angular_damping: 0.3
        body_density: 10.0
        body_friction: 0.3
        body_restitution: 0.5

        max_linear_speed: 100.0
        max_angular_speed: 2.0

        # Communication propeties
        communication_radius: 80.0# In mm, from each IR emitter (front, left, back, right)
        msg_success_rate:
            #type: static # Fixed msg success rate -- do not depend on the density, or message size
            #rate: 0.9
            type: dynamic # Msg success rate follows this formula:
                # "1 / (1 + (alpha * msg_size**beta * p_send**gamma * cluster_size**delta))"
            alpha: 0.000001
            beta: 3.0708
            gamma: 2.3234
            delta: 1.1897


### Parameters used to directly initialize global variables from the C code of the robots ###
parameters:
    # Configuration parameters for the run_and_tumble example
    run_duration_min: 1000
    run_duration_max: 5000
    tumble_duration_min: 100
    tumble_duration_max: 1100
    enable_backward_dir: true
\end{lstlisting}

Pogosim implements a collection of possible instantiable objects (not just Pogobots):
\begin{description}
    \item[pogobot] Baseline Pogobots.
    \item[pogobject] Pogobot head on a large body.
    \item[pogowall] Active walls, using IR LED strips attached to walls and objects.
    \item[membrane] Mobile pogowalls, implemented by joints attached through a string.
    \item[passive\_object] A simple object without any controller. Can still interact with the environment (e.g. be pushed by robots).
    \item[static\_light] A light spot of gradient over a given region of the arena.
\end{description}
See the examples provided in the main GIT repository of Pogosim to see how they can be created in configuration files.

%

\subsection{Robot communication}

\begin{figure}[htbp]
    \centering
    \includegraphics[width=1.0\textwidth]{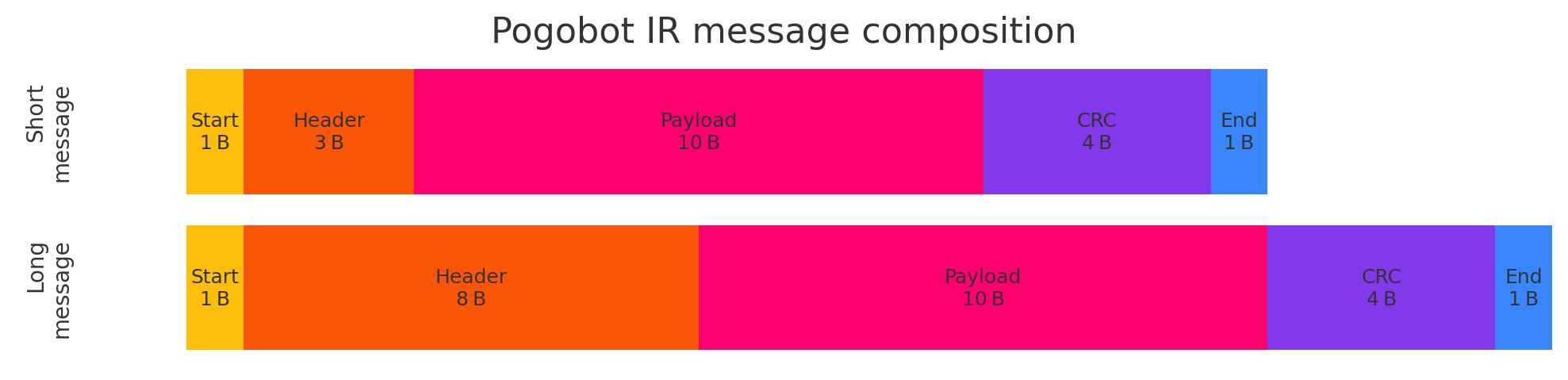}%
    \caption{Message layouts depending on their type: Short messages have a smaller header than Long messages.}
    \label{fig:messages_type}
\end{figure}

\begin{figure}[htbp]
    \centering
    \includegraphics[width=0.48\linewidth]{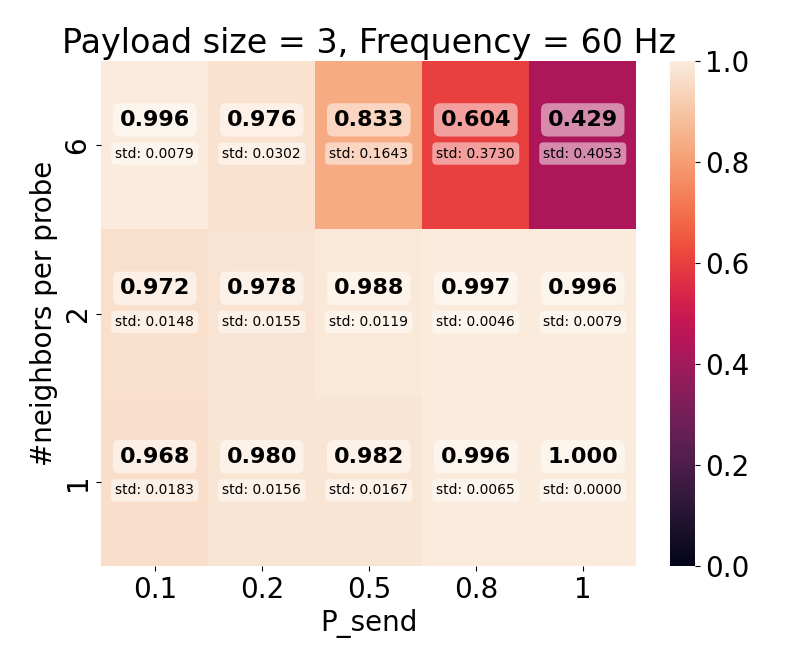}
    \includegraphics[width=0.48\linewidth]{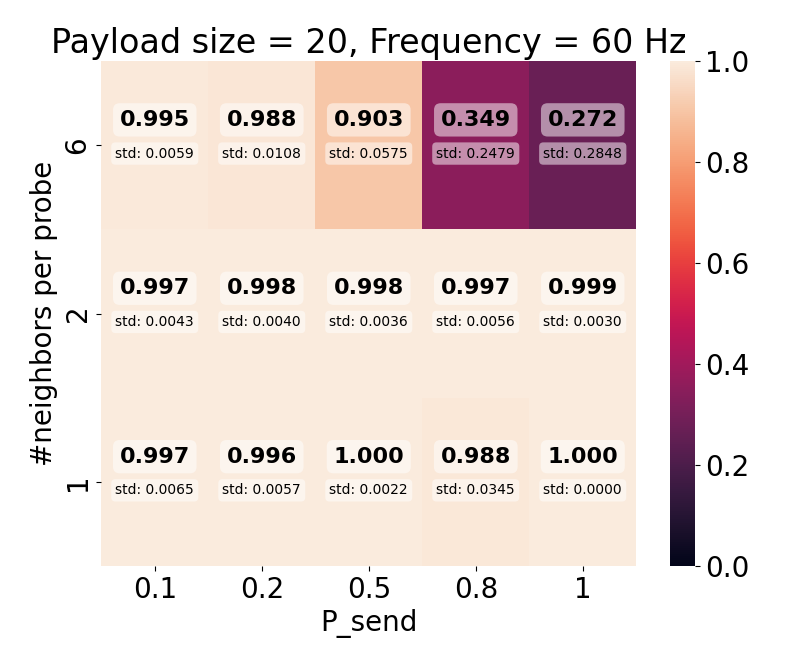}
    \includegraphics[width=0.48\linewidth]{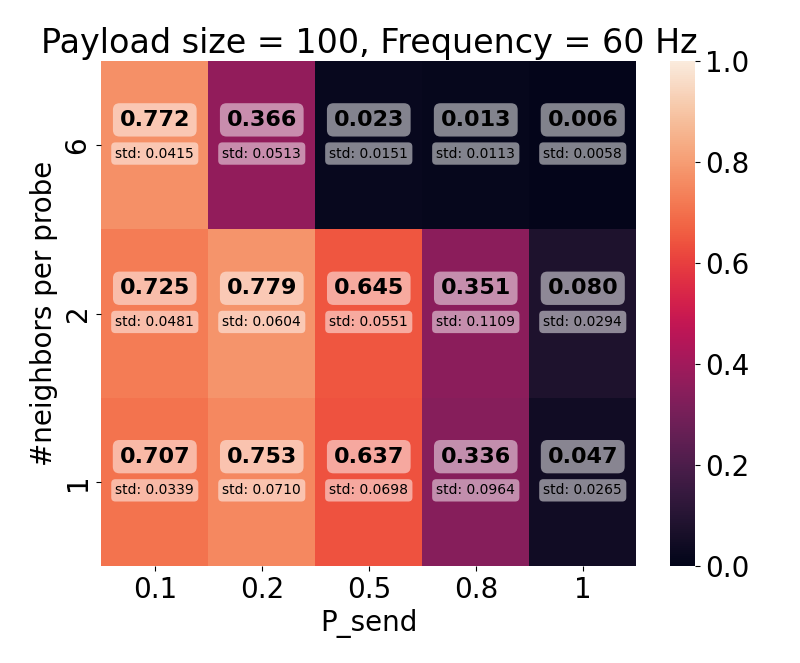}
    \caption{Probability of message reception success from a focal robot to other robots in its neighborhood across 10 runs and with a main loop frequency of 60Hz. We vary payload size (in short messages, so msg\_size = 3 + payload\_size), the number of robots in neighborhood, and $P_\texttt{send}$ the probability of sending a message at each pogotick.}
    \label{fig:reception_success}
\end{figure}

Pogobots communicate through IR with each other, with 4 IRs emitters distributed on the borders of the Pogobot head, and 8 IRs receptors. The communication API provides two types of messages that user codes can emit: short and long messages (Fig.~\ref{fig:messages_type}). Both types are implemented in Pogosim.

IR communication can be very noisy: Pogobots broadcast signals to their neighborhood, often resulting in message collision when they emit signals at the same time. The robot API already implements a CRC32 checksum mechanism to detect if a message was corrupted.

To reproduce the same behavior in simulations, we model the number of messages that are correctly received from a focal robot to each neighbor. Two models are available, as specified in configuration files.

Firstly, using a "static" model, the percentage of correctly received message follows a constant probability. It is specified as follows:
\begin{lstlisting}[basicstyle=\ttfamily\small,language=yaml]
# Communication properties
communication_radius: 80.0      # In mm, from each IR emitter.
msg_success_rate:
    # Fixed msg success rate -- do not depend on the density, or message size
    type: static
    rate: 0.9
\end{lstlisting}

Secondly, using a "dynamic" model that takes into account the density of the focal robot neighborhood, the size of the message, and the probability of each robot to send a message. 
We base our analysis on communication statistics obtained experimentally from real Pogobots (Fig.~\ref{fig:reception_success}).

We fit these data with the following model:
\begin{equation} \label{eq:eq_msg_sim}
P_{\texttt{recv}} = \frac{1}{\alpha + \beta \times {p_\texttt{send}}^{\gamma}
\times \texttt{cluster\_size}^\delta
\times  \texttt{exp}(\zeta \times \texttt{msg\_size})
+ \theta \times  {p_\texttt{send} \times \texttt{cluster\_size}}
}
%
\end{equation}
with $P_{\texttt{recv}}$ the probability of correctly receiving a message, $\texttt{msg\_size}$ the size of the messages in bytes (incl. both headers and payload), $p_\texttt{send}$ the probability at each pogoticks that a robot send a message, and $\texttt{cluster\_size}$ the number of robots in a communication neighborhood (including the focal robot, i.e. $\texttt{cluster\_size} = 1 + \texttt{nb\_neighbors}$). The model has 7 parameters: $\alpha$ (base denominator term), $\beta$ (scale factor for power terms), $\gamma$ (power exponent for $p_\texttt{send}$), $\delta$ (power exponent for $\texttt{cluster\_size}$), $\zeta$ (Exponential coefficient for $\texttt{msg\_size}$) and $\theta$ (Interaction coefficient).

The parameters of this model can be specified in the configuration:
\begin{lstlisting}[basicstyle=\ttfamily\small,language=yaml]
# Communication properties
communication_radius: 80.0      # In mm, from each IR emitter.
msg_success_rate:
    # Dynamic msg success rate -- depend on the density, and message size
    type: dynamic
    alpha: 1.03215183
    beta:  0.00073859
    gamma: 3.14782227
    delta: 3.52543753
    zeta:  0.05720136
    theta: 0.00100000
\end{lstlisting}
The default parameters were fitted on the experimental data of Fig.~\ref{fig:reception_success} using the CMA-ES black-box optimizer~\cite{hansen2003reducing}, aiming to maximize the $R^2$ score. The selected default parameters specified above correspond to $R^2 = 0.914975$, with $0.066285$ of overall MAE.

\begin{figure}[htbp]
    \centering
    \includegraphics[width=0.55\textwidth]{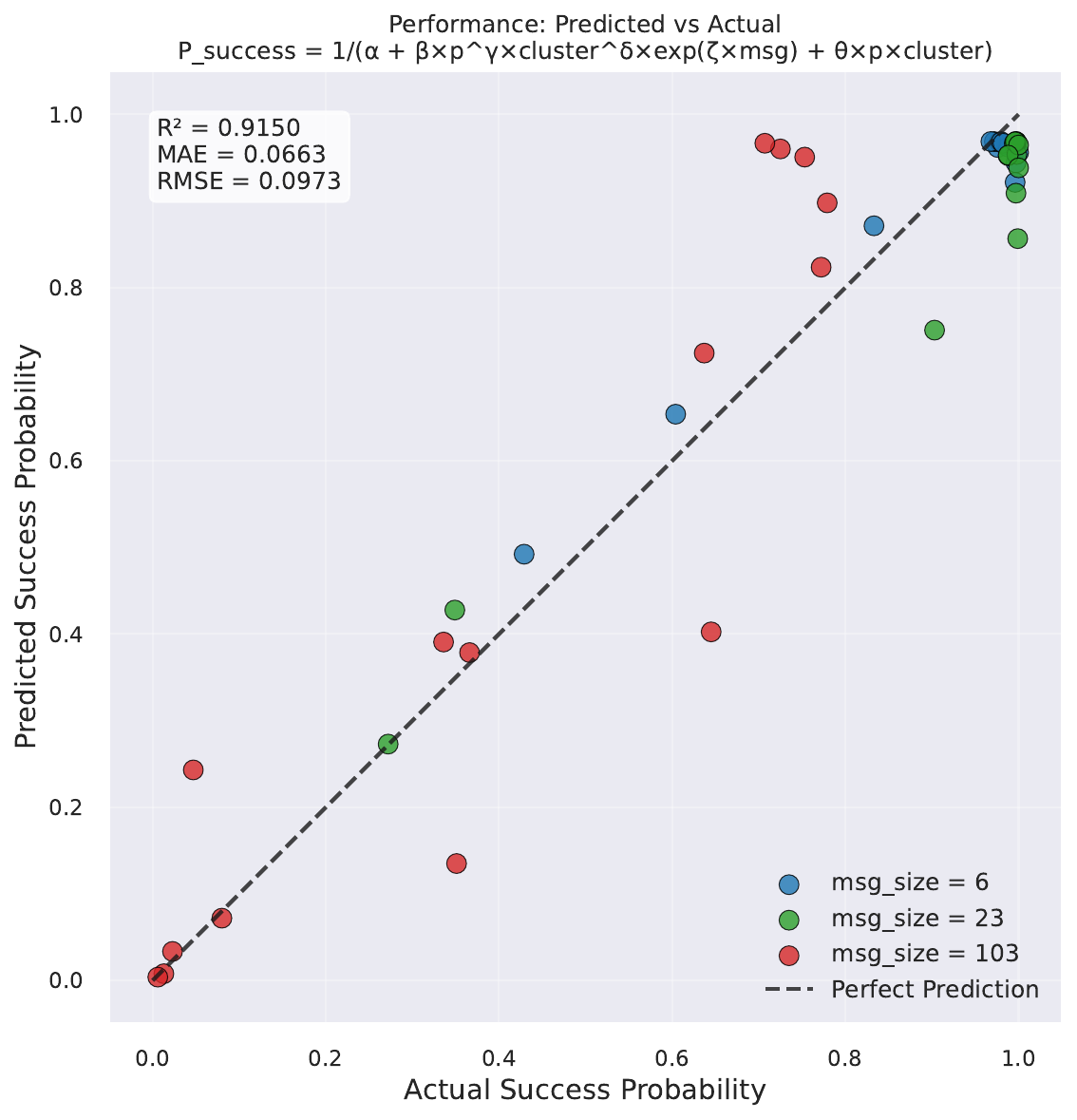}
    
    \includegraphics[width=0.45\linewidth]{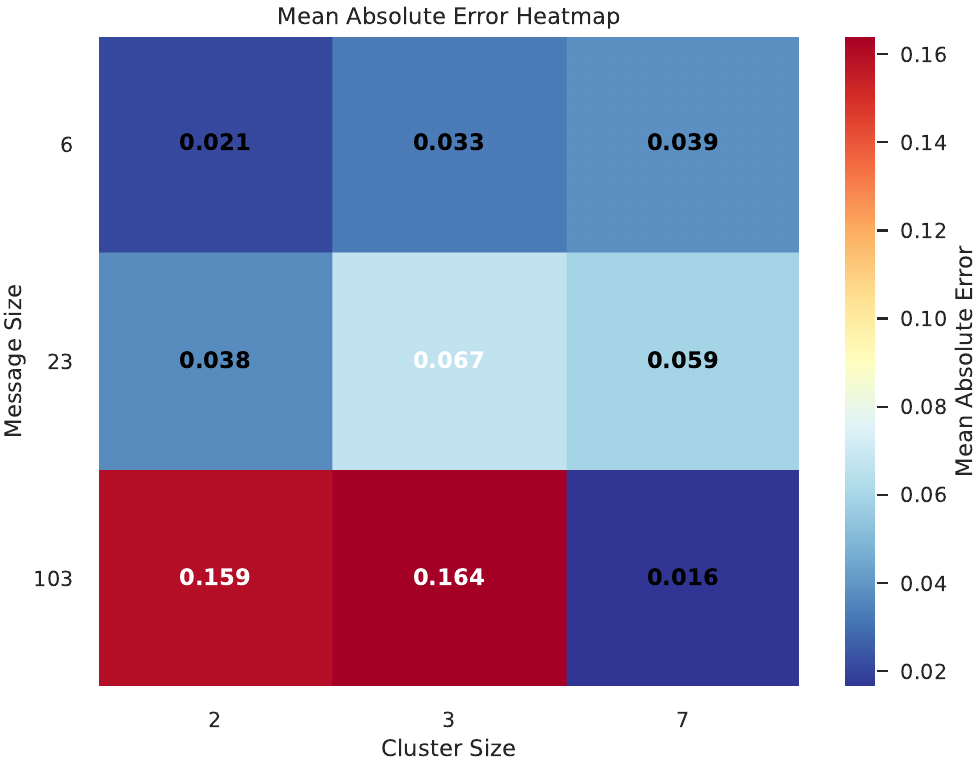}%
    \includegraphics[width=0.45\linewidth]{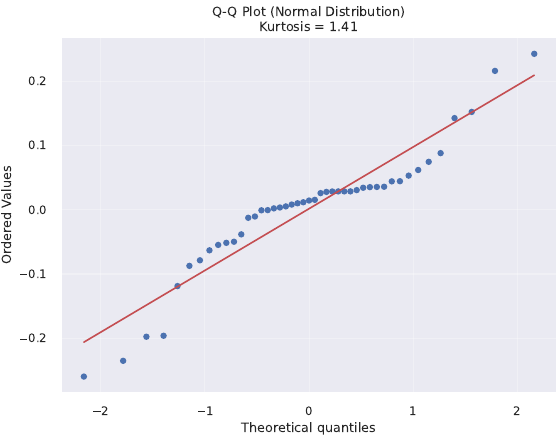}
    
    \includegraphics[width=0.99\textwidth]{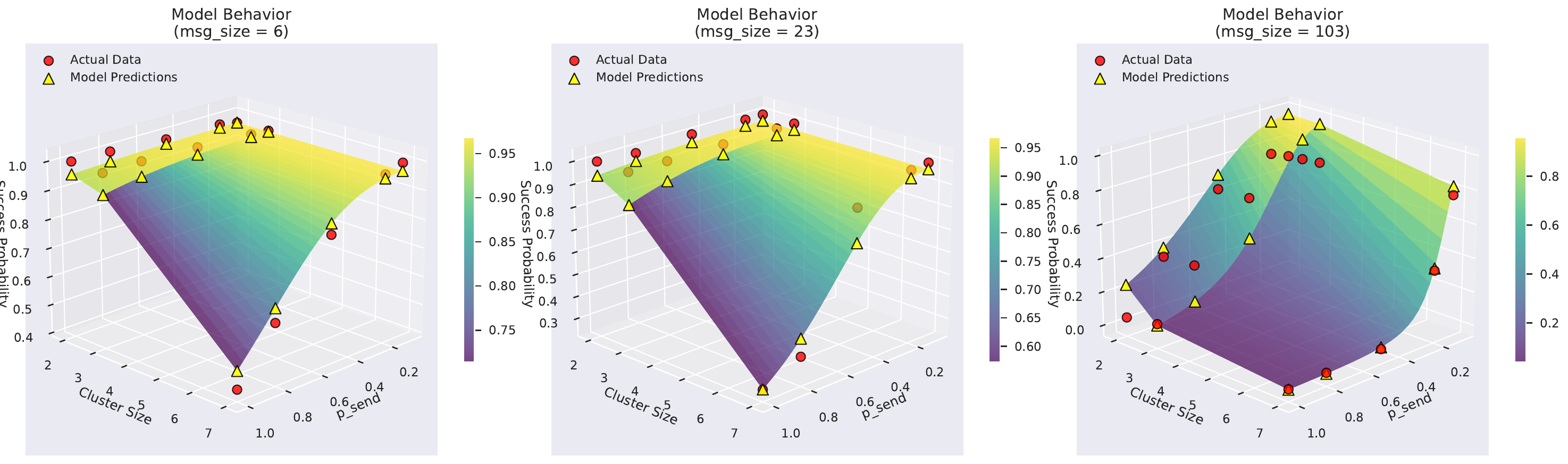}%
    \caption{\textbf{Top}: Predicted vs Actual Success Probabilities. Scatter plot comparing actual experimental success probabilities against model predictions. Points clustering near the perfect prediction line indicate good model fit, with some systematic deviation for large message sizes (red points) showing the remaining challenge in modeling high-payload communication scenarios.
    \textbf{Middle left}: Heatmap of mean absolute error by message size and cluster size, with numerical values showing worst performance for large messages in large clusters (0.112 MAE).
    \textbf{Middle right}: Q-Q plot against normal distribution, revealing a slight residuals deviation from normality at both extremes, suggesting overall good results with small difficulties at modeling outliers.
    \textbf{Bottom}: 3D Model Behavior Analysis Across Parameter Space.}
    \label{fig:comm_success_model_comparison}
\end{figure}

Figure~\ref{fig:comm_success_model_comparison} compares the "dynamic" model of message reception success with the experimental data of Fig.~\ref{fig:reception_success}: the model has a good overall fit, with some small (but acceptable) levels of errors in large message sizes.

\subsection{User interface}

\begin{figure}[htbp]
    \centering
    \includegraphics[height=10.0cm]{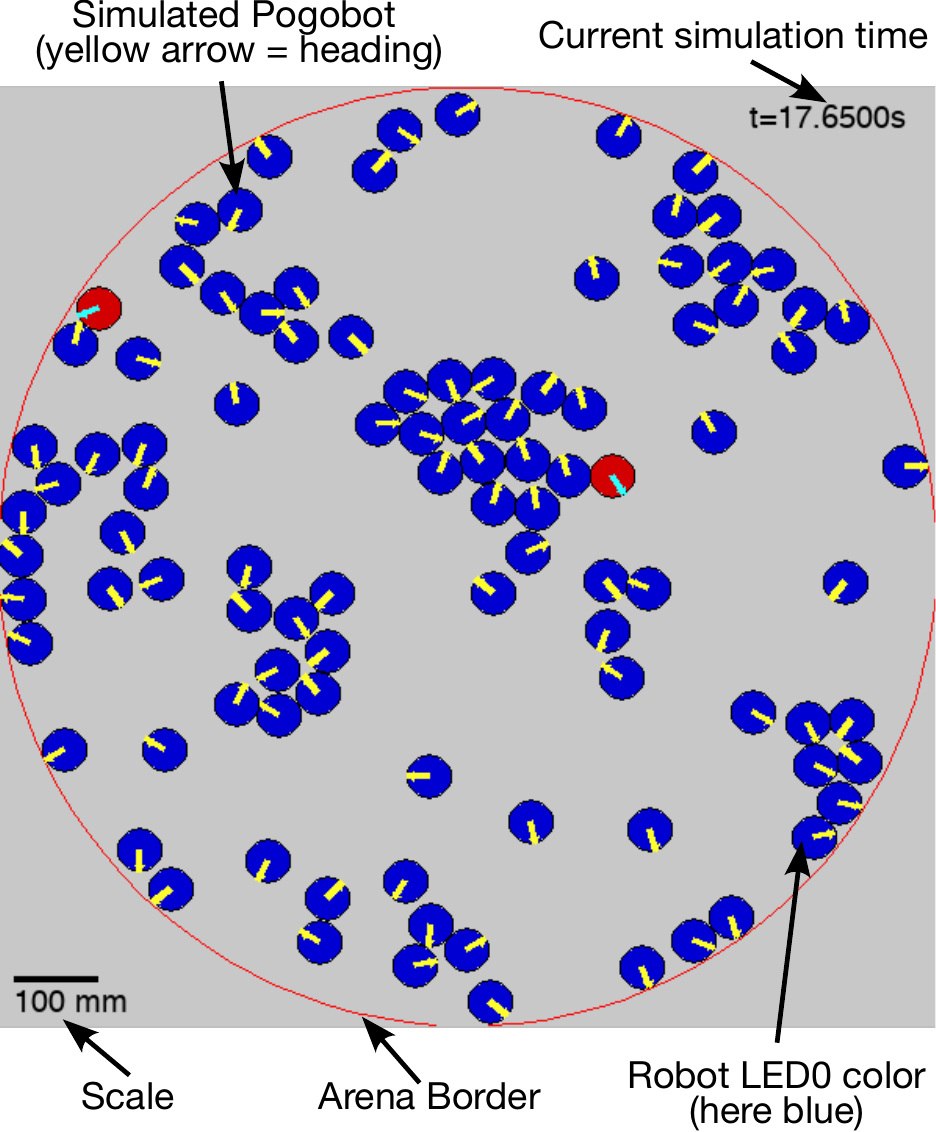}%
    \caption{Graphical User Interface of the Pogosim simulator, with a 100-robot setup.}
    \label{fig:gui}
\end{figure}

Pogosim uses the SDL2:~\cite{sdl2} library to implement the simulation visualization and GUI (Fig.~\ref{fig:gui}). The interface renders the simulation objects (robots, lights, passive objects, etc) and arena boundaries (walls), alongside a scale bar (bottom left corner) and current simulated time (top right corner).

\subsection{Serialization}
Pogosim serializes the robot pose and state at a given frequency (wrt the simulation time) into a compact Apache Arrow~\cite{arrow-manual} feather file. This format was selected (rather than plain CSV files) because it has a very good tradeoff speed vs file size. CSV files would become very large for simulations with even a relatively small number of agents (e.g. 100). Apache feather files in turn, will compress the result, allowing for very fast reads.

It is possible to specify the frequency of serialization in the configuration files:
\begin{lstlisting}[basicstyle=\ttfamily\scriptsize,language=yaml]
enable_data_logging: true
data_filename: "frames/data.feather"
console_filename: "frames/console.txt"
save_data_period: 1.0       # In s, or -1 to disable data export (see also "enable_data_logging")
\end{lstlisting}

The user can also specify custom columns that will be stored in the dataframe:
\begin{lstlisting}[basicstyle=\ttfamily\scriptsize]
#ifdef SIMULATOR
// Function called once by the simulator to specify user-defined data fields to add to the exported data files
void create_data_schema() {
    data_add_column_int32("stuff");
}

// Function called periodically by the simulator each time data is saved (cf config parameter "save_data_period" in seconds)
void export_data() {
    if (mydata->started) { // Only store data after the experiment has started
        enable_data_export(); // Enable data export this time
        data_set_value_int32("stuff", mydata->foobar);
    } else { // Disable data export this time
        disable_data_export();
    }
}
#endif

// Entrypoint of the program
int main(void) {
    pogobot_init();     // Initialization routine for the robots
    // Specify the user_init and user_step functions
    pogobot_start(user_init, user_step);

    // Specify the callback functions. Only called by the simulator.
    //  In particular, they serve to add data fields to the exported data files
    SET_CALLBACK(callback_create_data_schema, create_data_schema);  // Called once to specify the data format
    SET_CALLBACK(callback_export_data, export_data);                // Called at each period on each robot to register exported data
    return 0;
}
\end{lstlisting}

Afterwards, it is possible to open the feather file in Python using the Pandas library:
\begin{lstlisting}[basicstyle=\ttfamily\scriptsize,language=python]
import pandas as pd
df = pd.read_feather("frames/data.feather")
print(df)

       time robot_category  robot_id  pogobot_ticks         x         y       angle  
0      1.00          walls     65535             63  5.001000  5.001000    0.000000  
1      1.00      membranes     65534             63  6.824879  4.867349         NaN  
2      1.00         robots         0             63  4.038734  0.959281    1.953128  
3      1.00         robots         1             63  1.023770  8.115510    1.303922  
4      1.00         robots         2             63  1.965905  3.455247   -2.005039  
...     ...            ...       ...            ...       ...       ...         ...  
1423  14.13         robots        95            884  0.654096  9.729611   -0.386542  
1424  14.13         robots        96            884  8.011082  6.353422    1.812981  
1425  14.13         robots        97            884  4.807075  7.854455    1.177420  
1426  14.13         robots        98            884  4.021065  9.730083    1.626796  
1427  14.13         robots        99            883  9.031286  1.299344   -0.168437  

[1428 rows x 7 columns]
\end{lstlisting}

Another way would be to directly use the Pogosim Python scripts, which can be installed by:
\begin{lstlisting}[basicstyle=\ttfamily\scriptsize,language=bash]
pip3 install pogosim
\end{lstlisting}
Afterwards, you can use a function of this library to retrieve both the dataframe of results and the original configuration (YAML file) used to generate it:
\begin{lstlisting}[basicstyle=\ttfamily\scriptsize,language=python]
import pogosim.utils
df, metadata = pogosim.utils.load_dataframe("results/tmp/result.feather")
configuration = metadata.get("configuration", {})
print("### Here is the dataframe of results: ###\n", df)
print("\n### Here is the original configuration: ###")
import pprint
pprint.pprint(configuration)


### Here is the dataframe of results: ###
time robot_category  robot_id  pogobot_ticks           x           y  \
0    1.010000         robots         0             64  178.454254  178.462326   
1    1.010000         robots         1             64  231.984253  178.462326   
2    2.020000         robots         0            127  178.454254  178.462326   
3    2.020000         robots         1            127  231.984253  178.462326   
4    3.030000         robots         0            190  178.454254  178.462326   
..        ...            ...       ...            ...         ...         ...   
93  47.469999         robots         1           2967  231.984253  178.462326   
94  48.479999         robots         0           3030  178.454254  178.462326   
95  48.479999         robots         1           3030  231.984253  178.462326   
96  49.489999         robots         0           3094  178.454254  178.462326   
97  49.489999         robots         1           3094  231.984253  178.462326   

       angle  age  rgb_colors_index  run  
0  -2.233643    0                 0    0  
1  -0.972240    0                 0    0  
2  -2.233643    0                 0    0  
3  -0.972240    0                 0    0  
4  -2.233643    0                 0    0  
..       ...  ...               ...  ...  
93 -0.972240    2                 9    0  
94 -2.233643    2                 9    0  
95 -0.972240    2                 9    0  
96 -2.233643    2                 9    0  
97 -0.972240    2                 9    0  

[98 rows x 10 columns]

### Here is the original configuration: ###
{'GUI': True,
 'GUI_speed_up': 0.01,
 'arena_file': 'arenas/disk.csv',
 'arena_surface': '1.0e5',
 'arena_temperature': 25.0,
 'communication_ignore_occlusions': False,
 'console_filename': 'frames/console.txt',
 'data_filename': 'frames/data.feather',
 'delete_old_files': True,
 'enable_console_logging': True,
 'enable_data_logging': True,
 'frames_name': 'frames/f{:010.4f}.png',
 'initial_formation': 'disk',
 'objects': {'global_light': {'geometry': 'global',
                              'light_mode': 'static',
                              'photo_start_at': 1.0,
                              'photo_start_duration': 1.0,
                              'photo_start_value': 32767,
                              'type': 'static_light',
                              'value': 200},
             'robots': {'angular_noise_stddev': 0,
                        'body_angular_damping': 0.3,
                        'body_density': 10.0,
                        'body_friction': 0.3,
                        'body_linear_damping': 0.3,
                        'body_restitution': 0.5,
                        'communication_radius': 80.0,
                        'geometry': 'disk',
                        'linear_noise_stddev': 0,
                        'max_angular_speed': 2.0,
                        'max_linear_speed': 100.0,
                        'msg_success_rate': {'alpha': 1e-06,
                                             'beta': 3.0708,
                                             'delta': 1.1897,
                                             'gamma': 2.3234,
                                             'type': 'dynamic'},
                        'nb': 2,
                        'radius': 26.5,
                        'type': 'pogobot'}},
 'parameters': {'enable_backward_dir': True,
                'run_duration_max': 2000,
                'run_duration_min': 100,
                'tumble_duration_max': 1100,
                'tumble_duration_min': 100},
 'save_data_period': 1.0,
 'save_video_period': 1.0,
 'seed': 0,
 'show_communication_channels': False,
 'show_communication_channels_above_all': False,
 'show_lateral_LEDs': False,
 'show_light_levels': False,
 'simulation_time': 50.0,
 'time_step': 0.01,
 'window_height': 600,
 'window_width': 600}
\end{lstlisting}

\subsection{Examples}
Example codes are compiled every time you launch the "./build.sh" script, alongside the rest of the Pogosim code.

To launch examples code you can use the following commands:
\begin{lstlisting}[basicstyle=\ttfamily\scriptsize,language=bash]
# Hello world, just robots rotating left then right. The first robot prints "HELLO WORLD !" messages
./examples/helloworld/helloworld -c conf/simple.yaml

# A very simple implementation of the run-and-tumble algorithm for locomotion
./examples/run_and_tumble/run_and_tumble -c conf/simple.yaml

# A simple code to showcase the diffusion of information in a swarm.
#  Immobile robots by default (uncomment "MOVING_ROBOTS" to make then move)
./examples/hanabi/hanabi -c conf/simple.yaml

# An example showcasing phototaxis, with a fixed light spot in the middle of the arena
./examples/phototaxis/phototaxis -c conf/phototaxis.yaml

# An multi-controller example where robots can identify the presence of fixed walls (through Pogowalls)
#  or mobile walls (through membranes).
./examples/walls/walls -c conf/walls_and_membranes.yaml

# More complex example. "Simple" implementation of the SSR algorithm from https://arxiv.org/abs/2403.17147
# You can test it for a disk and annulus arena (see conf/ssr.yaml to change the arena).
./examples/ssr/ssr -c conf/ssr.yaml

# More complex run-and-tumble example, with two objectives: neighbor novelty,
#  and isolation avoidance (as a proxy to global coverage)
./examples/coverage_neighbors_novelty/coverage_neighbors_novelty -c conf/coverage_neighbors_novelty.yaml

# A run-and-tumble example showing how to retrieve IMU information (gyroscope, accelerometer, temperature sensor)
./examples/IMU/IMU -c conf/simple.yaml

# Canonical example of the push-sum gossip algorithm.
./examples/push_sum/push_sum -c conf/simple.yaml

# Showcases a Kuramoto-style moving oscillators swarm achieving synchronization.
#  The robots move according to a run-and-tumble algorithm.
./examples/moving_oscillators/moving_oscillators -c conf/simple.yaml

# Robot estimate their X,Y position using two rotating lighthouse inspired by the **Valve SteamVR** tracking system,
#  often used with drone localization.
./examples/lighthouse_localization/lighthouse_localization -c conf/lighthouse.yaml
\end{lstlisting}

The examples "run\_and\_tumble", "hanabi", "phototaxis", "walls" and "SSR" are all seen with different arenas in Fig.~\ref{fig:gallery}.

\subsection{Running simulations in batch and across configurations}
We provide Python scripts that can launch several runs of Pogosim in parallel, and compile the results from all runs into a single dataframe.

To install it, use the following command:
\begin{lstlisting}[basicstyle=\ttfamily\small,language=bash]
pip install pogosim
\end{lstlisting}

Or, just you want to compile it yourself:
\begin{lstlisting}[basicstyle=\ttfamily\small,language=bash]
cd scripts
./setup.py sdist bdist_wheel
pip install -U .
cd ..
\end{lstlisting}

Afterwards, you can use the pogobatch script to launch several runs of simulation in parallel (or in a cluster), with a given configuration:
\begin{lstlisting}[basicstyle=\ttfamily\small,language=bash]
pogobatch -c conf/test.yaml -S ./examples/hanabi/hanabi -r 10 -t tmp -o results
\end{lstlisting}

This command with launch 10 runs of the Hanabi example using configuration file conf/test.yaml. Temporary files of the runs will be stored in the "tmp" directory.
After all runs are completed, the script will compile a dataframe of all results and save it into "results/result.feather". It can then be opened as described in previous section. An additional column "run" is added to the dataframe to distinguish results from the different runs.

It is also possible to launch the pogobatch script on several variations of a given configuration, e.g. with a list of different numbers of robots or arena. The list of possibly configuration combination is specified in the configuration file, by adding a subkey "batch\_options" with the list of possible values.
E.g.:
\begin{lstlisting}[basicstyle=\ttfamily\small,language=yaml]
arena_file:        
    # Test the results on two arenas
    batch_options: ["arenas/disk.csv", "arenas/arena8.csv"]
    # OPTIONAL: Value to use for "arena_file"
    #  when this configuration is used directly by the simulator, not pogobatch
    default_option: arenas/disk.csv
objects:
    robots:
        type: pogobot       # Category type pertaining to Pogobots
        nb:                 # Number of objects (Pogobots) in this category
            # Test the results on three different swarm sizes
            batch_options: [100, 200]
            # OPTIONAL: Value to use for "objects.robots.nb"
            #  when this configuration is used directly by the simulator, not pogobatch
            default_option: arenas/disk.csv
        geometry: disk                  # Pogobots are always disk-shaped
        radius: 26.5                    # In mm


# Format of the generated dataframes, one for each configuration
result_filename_format: "result_{objects.robots.nb}.feather"

# List of new columns to add in the generated dataframes
result_new_columns: ["arena_file"]
\end{lstlisting}
These configuration entries specify that either 100 or 200 robots should be considered, on arenas "disk" and "8", resulting in 4 possibly configurations. The configuration entry "result\_filename\_format" corresponds to the name of a given configuration combination.
See "conf/batch/test.yaml" for a complete example. The entry "result\_new\_columns" indicates which columns (and associated configurations) are *stored* inside feather files as additional columns.

You can use pogobatch script on this compounded configuration file to launch several runs on each configuration combination:
\begin{lstlisting}[basicstyle=\ttfamily\small,language=bash]
pogobatch -c conf/batch/test.yaml -S ./examples/hanabi/hanabi -r 10 -t tmp -o results

Found 6 combination(s) to run.
Task: Config file /home/syemn/data/prj/pogosim/tmp/combo_kdmvxpzf.yaml -> Output: results/result_50.feather
Task: Config file /home/syemn/data/prj/pogosim/tmp/combo_cqfk3lrl.yaml -> Output: results/result_100.feather
Task: Config file /home/syemn/data/prj/pogosim/tmp/combo_ckx7t160.yaml -> Output: results/result_150.feather
Task: Config file /home/syemn/data/prj/pogosim/tmp/combo_401kcmam.yaml -> Output: results/result_50.feather
Task: Config file /home/syemn/data/prj/pogosim/tmp/combo_wiilbu4e.yaml -> Output: results/result_100.feather
Task: Config file /home/syemn/data/prj/pogosim/tmp/combo_q3yxls9z.yaml -> Output: results/result_150.feather
Removed stale result file: results/result_50.feather
Removed stale result file: results/result_150.feather
Removed stale result file: results/result_100.feather
Launch -> tmp tmp/run_c03834a11a87406384efdcf8d2376dd4.feather  (will merge into results/result_50.feather)
Combined data saved to tmp/run_c03834a11a87406384efdcf8d2376dd4.feather
Created results/result_50.feather with 49500 rows
Launch -> tmp tmp/run_2628efb844fc4e2e83da56f7e98e8084.feather  (will merge into results/result_100.feather)
Combined data saved to tmp/run_2628efb844fc4e2e83da56f7e98e8084.feather
Created results/result_100.feather with 99000 rows
Launch -> tmp tmp/run_3c868b57dff0483c920ebf656b8c2eff.feather  (will merge into results/result_150.feather)
Combined data saved to tmp/run_3c868b57dff0483c920ebf656b8c2eff.feather
Created results/result_150.feather with 148500 rows
Launch -> tmp tmp/run_9e26f0240f8a4105aaf6851e5614ab93.feather  (will merge into results/result_50.feather)
Combined data saved to tmp/run_9e26f0240f8a4105aaf6851e5614ab93.feather
Appended 49500 rows to results/result_50.feather
Launch -> tmp tmp/run_143db3464fdd456c8cac379f621ae474.feather  (will merge into results/result_100.feather)
Combined data saved to tmp/run_143db3464fdd456c8cac379f621ae474.feather
Appended 99000 rows to results/result_100.feather
Launch -> tmp tmp/run_84b9da00ad624d788d7bcce6c301f8ca.feather  (will merge into results/result_150.feather)
Combined data saved to tmp/run_84b9da00ad624d788d7bcce6c301f8ca.feather
Appended 148500 rows to results/result_150.feather
Batch run completed. Generated output files:
 - results/result_50.feather
 - results/result_100.feather
 - results/result_150.feather
 - results/result_50.feather
 - results/result_100.feather
 - results/result_150.feather
\end{lstlisting}

If you want to implement more complex deployment behaviors, you can write your own Python scripts and extend the class "pogosim.pogobatch.PogobotBatchRunner".

\subsection{Offline parameter optimization}
The same "pogosim" python package as described in the previous section also includes a program to perform offline parameter optimization of a given Pogosim user code: "Pogoptim".

To use Pogoptim, the user can provide a configuration file with the list of parameters to be optimized and their domain (min, max values), and a python code with an objective function (e.g. motility dynamics like Mean Squared Displacement, Alignment, etc; or performance metrics for a given task). Pogoptim will then test a large number of parameter sets to maximize the objective function. 

Parameters to be optimized can just be specified in the configuration files. E.g.:
\begin{lstlisting}[basicstyle=\ttfamily\small,language=yaml]
# Parameters used to directly initialize variables from the C code of the robots
parameters:
    # Configuration parameters for the run_and_tumble example
    run_duration_min: 0
    run_duration_max:
        optimization_domain: {type: int, min: 10, max: 1000, init: 50}
    tumble_duration_min: 100
    tumble_duration_max: 1100
\end{lstlisting}
We implement 3 types of optimizable values: "int", "float" and "categorical".
    
We currently provide 3 optimization methods (more will be implemented in subsequent versions):
\begin{description}
    \item[random] Just perform random search, using uniform distribution in the parameters domain
    \item[cmaes] Use the CMA-ES~\cite{hansen2003reducing} black evolution-strategy algorithm. Useful for complex scenarios with ill-defined, non-linear objective landscapes
    \item[mapelites] Use the MAP-Elites~\cite{mouret2015illuminating} Quality-Diversity algorithm~\cite{pugh2016quality}, implemented in Python by the QDpy library~\cite{qdpy}. Useful to obtain a collection of diverse and high-performing solutions across Feature-Descriptors axes.
\end{description}

The objective function (for the cmaes and mapelites cases) and the feature descriptor function (for the mapelites case) can be defined in a python script file as follows:
\begin{lstlisting}[basicstyle=\ttfamily\small,language=python]
def objective_fn(df: pd.DataFrame) -> float:
    """Return the mean of per-agent MSD across *all* runs and arenas."""
    msd_df = compute_msd_per_agent(df)
    if msd_df.empty:
        logger.warning("Default MSD objective: empty input produced no MSD values; returning -inf")
        return float('-inf')
    return float(msd_df['MSD'].mean())

def fd_fn(df: pd.DataFrame) -> np.ndarray:
    """Return default 2-D features for QD: (max MSD, stddev MSD)."""
    msd_df = compute_msd_per_agent(df)
    if msd_df.empty:
        return np.array([0.0, 0.0], dtype=float)
    vals = np.asarray(msd_df["MSD"], dtype=float)
    return np.array([float(np.max(vals)), float(np.std(vals, ddof=0))], dtype=float)
\end{lstlisting}
If no objective/FD functions are provided, Pogoptim will use the previous functions by default (and print out a warning).

Here is how to launch pogoptim:
\begin{lstlisting}[basicstyle=\ttfamily\scriptsize,language=bash]
./scripts/pogosim/pogoptim.py -c conf/optim/simple.yaml -S ./examples/run_and_tumble/run_and_tumble \
  -r 5 --max-evals 10 --optimizer cmaes --objective obj.py

[15:35:47] [WARNING] No objective script provided; DEFAULT fitness = mean MSD (features default to [polar order, straightness]).
[15:35:47] [INFO] Starting optimization: CMAES | dim=1 | max_evals=10 (normalized space)
[15:35:57] [INFO] gen 001: pop=4  f[best/mean/min]=[80143.3/60560.4/44425.7]  best_so_far=80143.3
[15:36:07] [INFO] gen 002: pop=4  f[best/mean/min]=[102601/75686.4/61625.2]  best_so_far=102601
[15:36:13] [INFO] gen 003: pop=2  f[best/mean/min]=[109715/109102/108489]  best_so_far=109715
[15:36:13] [INFO] Done. Best fitness: 109715
[15:36:13] [INFO] Best values: {"parameters.run_duration_max": 1000}
\end{lstlisting}
with "-S" used to specify the simulation binary,  "-r 5" the number of runs per evaluation, "--max-evals" the evaluation budget, "--optimizer" the name of the optimizer and "--objective" used to specify the python script containing the objective and FD functions.


\section{User code and code portability}

\begin{table}[h!]
\centering
\footnotesize
\begin{tabular}{p{6cm} p{5.5cm} p{6.0cm}}
\hline
Baseline Pogobot API & Code adapted for Pogosim & Comments \\
\hline
\lstinline[basicstyle=\ttfamily\tiny]|#include "pogobot.h"| & \lstinline[basicstyle=\ttfamily\tiny]|#include "pogobase.h"| & The "pogobase.h" header file automatically includes routines either for a simulation or for physical Pogobots. \\

\hline
\begin{minipage}[t]{\linewidth}
\vspace{-\baselineskip}
\begin{lstlisting}[basicstyle=\ttfamily\tiny]
uint16_t iteration;
uint8_t age;
uint8_t rgb_colors_index;
\end{lstlisting}
\end{minipage}
&
\begin{minipage}[t]{\linewidth}
\vspace{-\baselineskip}
\begin{lstlisting}[basicstyle=\ttfamily\tiny]
typedef struct {
    uint16_t iteration;
    uint16_t age;
    uint8_t rgb_colors_index;
} USERDATA;
\end{lstlisting}
\end{minipage}
& "Global" variables should be inserted within the USERDATA struct. In simulation, don't declare non-const global variables outside this struct, elsewise they will be shared among all agents (only make sense for global configuration parameters).\\

\hline
&
\begin{minipage}[t]{\linewidth}
\vspace{-\baselineskip}
\begin{lstlisting}[basicstyle=\ttfamily\tiny]
// Call this macro in the same file
//  (.h or .c) as the declaration of USERDATA
DECLARE_USERDATA(USERDATA);
// Don't forget to call this macro in the
// main .c file of your project (only once!)
REGISTER_USERDATA(USERDATA);
\end{lstlisting}
\end{minipage}
& These macros define the pointer mydata (same idea as in the Kilombo simulator~\cite{jansson2015kilombo}).  Now, members of the USERDATA struct can be accessed through \lstinline|mydata->MEMBER|. On real robots, the compiler will automatically optimize the code to access member variables as if they were true globals.\\

\hline
\begin{minipage}[t]{\linewidth}
\vspace{-\baselineskip}
\begin{lstlisting}[basicstyle=\ttfamily\tiny]
#define FQCY 60 // 60 Hz
// Entrypoint of the program
int main(void) {
  // Initialization routine
  pogobot_init();
  time_reference_t w;

  // Main loop
  uint32_t pogobot_ticks = 0;
  for (;;) {
    if (pogobot_ticks % 1000 == 0
        && pogobot_helper_getid() == 0) {
      printf("HELLO WORLD!\n"));
    }
    mydata->data_foo[0] = 42;
    ms =  pogobot_stopwatch_get_elapsed_microseconds(&w);
    // Wait for next step
    if (ms < 1000000 / FQCY) {
      msleep( (1000000 / FQCY - microseconds ) / 1000 );
    }
    pogobot_ticks++;
  }
  return 0;
}
\end{lstlisting}
\end{minipage}
&
\begin{minipage}[t]{\linewidth}
\vspace{-\baselineskip}
\begin{lstlisting}[basicstyle=\ttfamily\tiny]
// Init function. Called once at the beginning of the program.
void user_init(void) {
  // Call the 'user_step' function 60 times per second
  main_loop_hz = 60;      
}

// Step function.
// Called continuously at each step of the main loop
void user_step(void) {
  // Only print messages for robot 0
  if (pogobot_ticks % 1000 == 0
      && pogobot_helper_getid() == 0) {
    printf("HELLO WORLD!\n");
  }
  mydata->data_foo[0] = 42;
}

// Entrypoint of the program
int main(void) {
  pogobot_init(); // Initialization routine for the robots
  // Specify the user_init and user_step functions
  pogobot_start(user_init, user_step);
  return 0;
}
\end{lstlisting}
\end{minipage}
& Pogosim introduces the \lstinline[basicstyle=\ttfamily\scriptsize]|pogostart(user_init, user_step)| function and split the original \lstinline|main(void)| function architecture into \lstinline|user_init(void)| and \lstinline|user_step(void)| functions.   \\

\hline
\begin{minipage}[t]{\linewidth}
\vspace{-\baselineskip}
\begin{lstlisting}[basicstyle=\ttfamily\tiny]
#define FQCY 60 // 60 Hz
typedef struct __attribute__((__packed__)) {
  uint16_t sender_id; // Unique identifier of the sender.
} heartbeat_t;
// Size of the message in bytes
#define MSG_SIZE ((uint16_t)sizeof(heartbeat_t))
uint16_t p_send_per_step = 50;

// Entrypoint of the program
int main(void) {
  // Initialization routine
  pogobot_init();
  time_reference_t w;

  // Main loop
  uint32_t pogobot_ticks = 0;
  for (;;) {
    // Process messages
    pogobot_infrared_update();
    if (pogobot_infrared_message_available()) {
      // Recover the next message
      while (pogobot_infrared_message_available()) {
        message_t mr;
        pogobot_infrared_recover_next_message(&mr);
        heartbeat_t const *hb = (heartbeat_t const *)mr.payload;
        uint16_t sender = hb->sender_id;
        printf("Received message from neighbor:%u\n", sender);
      }
    }
    pogobot_infrared_clear_message_queue();

    // Send message
    if (rand()%100 <= p_send_per_step) {
      heartbeat_t hb = { .sender_id = pogobot_helper_getid() };
      pogobot_infrared_sendShortMessage_omni(
        (uint8_t *)&hb, MSG_SIZE);
    }

    // Wait for next step
    ms = pogobot_stopwatch_get_elapsed_microseconds(&w);
    if (ms < 1000000 / FQCY) {
      msleep( (1000000 / FQCY - microseconds ) / 1000 );
    }
    pogobot_ticks++;
  }
  return 0;
}
\end{lstlisting}
\end{minipage}
&
\begin{minipage}[t]{\linewidth}
\vspace{-\baselineskip}
\begin{lstlisting}[basicstyle=\ttfamily\tiny]
typedef struct __attribute__((__packed__)) {
  uint16_t sender_id; // Unique identifier of the sender.
} heartbeat_t;
// Size of the message in bytes
#define MSG_SIZE ((uint16_t)sizeof(heartbeat_t))

// Called by the pogobot main loop before 'user_step'
bool send_message(void) {
  heartbeat_t hb = { .sender_id = pogobot_helper_getid() };
  pogobot_infrared_sendShortMessage_omni(
    (uint8_t *)&hb, MSG_SIZE);
  return true; // Return true if a message was sent
}

// Called by the pogobot main loop before 'user_step',
//   every time there is a new message
void process_message(message_t *mr) {
  if (mr->header.payload_length < MSG_SIZE) {
    return; // Not a heartbeat
  }
  heartbeat_t const *hb = (heartbeat_t const *)mr->payload;
  uint16_t sender = hb->sender_id;
  printf("Received message from neighbor:%u\n", sender);
}

void user_init(void) {
  // Set msg sending/processing freq
  main_loop_hz = 60;
  max_nb_processed_msg_per_tick = 3;
  percent_msgs_sent_per_ticks = 50;
  // Specify functions to send/transmit messages
  msg_rx_fn = process_message;
  msg_tx_fn = send_message;
}

\end{lstlisting}
\end{minipage}
& Pogosim introduce the \lstinline|send_message(void)| and \lstinline|process_message(message_t *mr)| functions that will be called at every \lstinline|pogoticks| (i.e. at every instance of the main loop) to handle message sending and processing. \\

\hline
\end{tabular}
\caption{List of differences in code routines between the baseline Pogobot API~\cite{loi2025pobogot} and the code used in the Pogosim simulator. The adapted code can still be compiled as user program of Pogobot robots without any modification, in addition to being usable by Pogosim.}
\label{tab:code_sim_vs_raw}
\end{table}

Pogosim integrates a lightweight ($<200$ C lines) interface layer to ease the simulator design and to allow the exact same code to be used in both simulation and real Pogobot experiments. Through an extensive use of macros and compilation options, this interface layer operates as syntactic sugar, with no impact on performance. Several routines are different compared to the baseline API, as listed in Table~\ref{tab:code_sim_vs_raw}.

In particular, variables describing bot internal states cannot remain global in Pogosim, elsewise all robots would have the same internal state (cf. Sec.~\ref{sec:design}). Taking inspiration from Kilombo~\cite{jansson2015kilombo}, we provide code routines and macros that creates a USERDATA structure to contain the bot internal states.

Another major difference from the baseline Pogobot API is that the infinite main loop of the program is not allowed to stay in the main function. Instead, we provide a function "pogobot\_start" that is called in the main function to specify an initialization function (e.g. "user\_init") executed before the main loop, and a step function (e.g. "user\_step") that is executed at every instance of the main loop. In simulation, Pogosim sequentially launches the latter on each robot, with their respective internal states (e.g. USERDATA variables) promoted as globals. 



%
%


\clearpage
\section{Discussions and Conclusion}

We presented a new swarm robotics simulator, targeting Pogobot robots. Pogosim is design to reproduce Pogobot hardware and APIs so that the exact same user code can be used both in simulation and to control real Pogobots in experiments. We proposed a collection of user code examples with a diversity of scopes, goals and complexity, showcasing all available capabilities on these robots.

Pogosim can scale to a large ($> 1000$) number of robots. It also integrates baseline scripts to 1) launch a large number of simulation runs across given parameters; 2) find user code parameters that optimize a given objective function; 3) analyze the dynamics of the swarm (motility, communication graph, etc). 

Pogosim will remain in active development, co-evolving with the real Pogobot robot and integrating new features as soon as they are added to the Pogobot hardware.


\section*{Acknowledgments}

This work was supported by the SSR project funded by the Agence Nationale pour la Recherche under Grant No ANR-24-CE33-7791. We thank Keivan Amini, Alessia Loi, Salman Houdaibi, Lara Polachini and Nathanael Aubert-Kato for meaningful comments and for regularly testing Pogosim during its development. We thank Alexandre Guerre and Leo Dupuyl for Pogobot hardware support.

\nolinenumbers

\bibliography{biblio.bib}

\bibliographystyle{unsrt}

\end{document}